\documentclass[sigplan,screen]{acmart}

\AtBeginDocument{%
  }

\setcopyright{acmlicensed}
\copyrightyear{2018}
\acmYear{2018}
\acmConference[]{}{ }{}
\acmISBN{978-1-4503-XXXX-X/2018/06}

\RequirePackage{subcaption}

\usepackage{fancyhdr}  % 使用fancyhdr包自定义页眉
\usepackage{pdfcomment}
\usepackage{xcolor}

\begin{document}

\title{Hear-Your-Click: Interactive Object-Specific Video-to-Audio Generation}

\author{Yingshan Liang}
\affiliation{%
  \institution{Shenzhen International Graduate School, Tsinghua University}
  \state{Shenzhen}
  \country{China}
}
\email{liangys23@mails.tsinghua.edu.cn}

\author{Keyu Fan}
\affiliation{%
  \institution{Shenzhen International Graduate School, Tsinghua University}
\state{Shenzhen}
\country{China}
}
\email{fky23@mails.tsinghua.edu.cn}

\author{Zhicheng Du}
\affiliation{%
	\institution{Shenzhen International Graduate School, Tsinghua University}
	\state{Shenzhen}
	\country{China}
}
\email{duzc24@mails.tsinghua.edu.cn}

\author{Yiran Wang}
\affiliation{%
	\institution{Shenzhen International Graduate School, Tsinghua University}
	\state{Shenzhen}
	\country{China}
}
\email{wangyr23@mails.tsinghua.edu.cn}

\author{Qingyang Shi}
\affiliation{%
	\institution{Shenzhen International Graduate School, Tsinghua University}
	\state{Shenzhen}
	\country{China}
}
\email{shiqy23@mails.tsinghua.edu.cn}

\author{Xinyu Zhang}
\affiliation{%
	\institution{Shenzhen International Graduate School, Tsinghua University}
	\state{Shenzhen}
	\country{China}
}
\email{z-xy23@mails.tsinghua.edu.cn}

\author{Jiasheng Lu}
\affiliation{%
  \institution{Huawei Technologies Co., Ltd.}
  \city{Shenzhen}
  \country{China}}
\email{lujiasheng2@huawei.com}

\renewcommand{\shortauthors}{Liang et al.}

\begin{abstract}
	Video-to-audio (V2A) generation shows great potential in fields such as film production. Despite significant advances, current V2A methods relying on global video information struggle with complex scenes and generating audio tailored to specific objects. To address these limitations, we introduce Hear-Your-Click, an interactive V2A framework enabling users to generate sounds for specific objects by clicking on the frame. To achieve this, we propose Object-aware Contrastive Audio-Visual Fine-tuning (OCAV) with a Mask-guided Visual Encoder (MVE) to obtain object-level visual features aligned with audio. Furthermore, we tailor two data augmentation strategies, Random Video Stitching (RVS) and Mask-guided Loudness Modulation (MLM), to enhance the model's sensitivity to segmented objects. To measure audio-visual correspondence, we designed a new evaluation metric, the CAV score. Extensive experiments demonstrate that our framework offers more precise control and improves generation performance across various metrics. \textit{Project Page: \url{https://github.com/SynapGrid/Hear-Your-Click}}
\end{abstract}



\settopmatter{printacmref=false}
\renewcommand\footnotetextcopyrightpermission[1]{}

\begin{teaserfigure}
	\centering
	\includegraphics[width=0.6\textwidth]{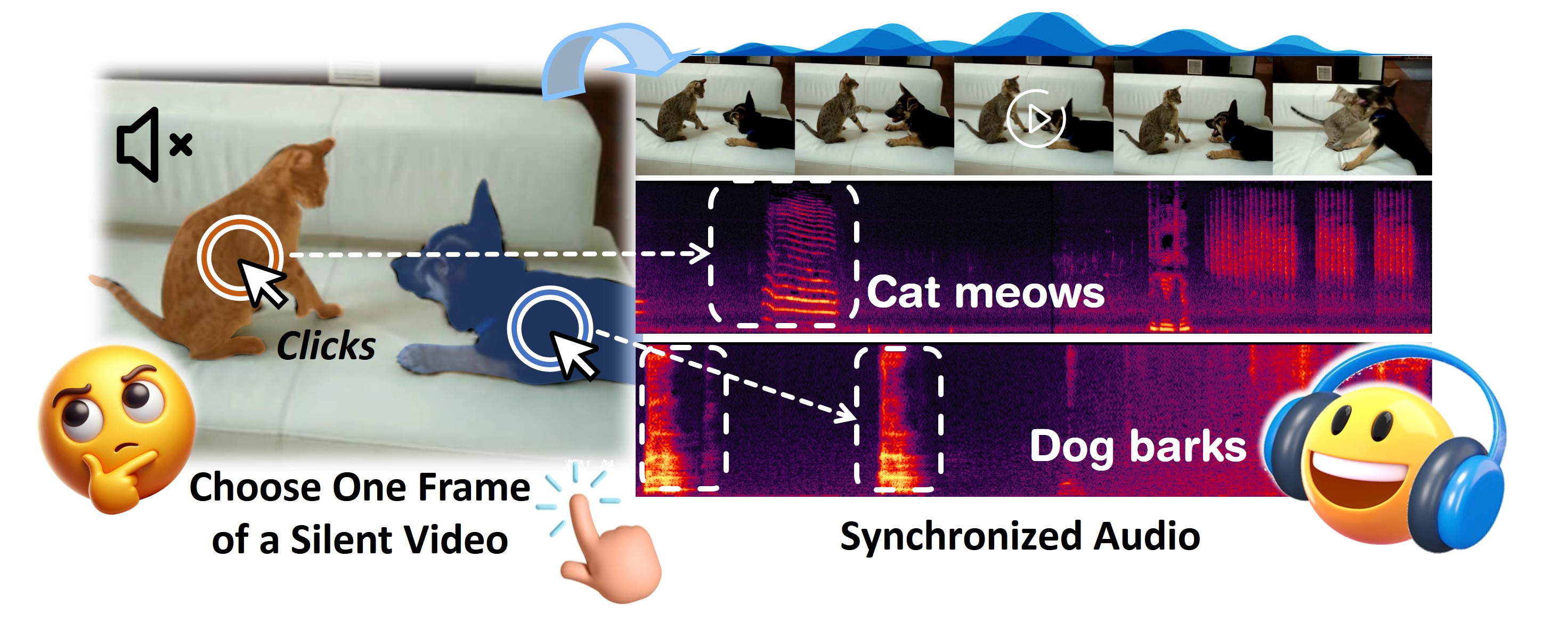}
	\caption{Overview of the Hear-Your-Click system. After users select a single frame and interactively choose specific objects or areas, Hear-Your-Click generates the corresponding audio. This user-friendly approach empowers individuals to customize the audio according to their preferences and interests.}
	\label{fig:overview}
\end{teaserfigure}

\maketitle
\raggedbottom
\section{Introduction}
\label{sec:intro}

Video-to-Audio (V2A) generation has great potential for applications in domains such as film production, social media, and accessibility services. Recent research in the field has yielded significant advances, particularly with diffusion models demonstrating remarkable performance in cross-modal generation tasks \cite{2023dreambooth,2023noise2music,2024fast,2024follow-your-click,2024brushnet}.

Despite considerable advances, several challenges remain in the field of V2A generation. One key challenge is comprehending intricate semantic information within videos. Videos often contain multiple objects and actions, requiring V2A models to generate appropriate sound effects for each object and their interactions. Consequently, most V2A models struggle to handle such rich semantic information.

Another challenge is the lack of fine-grained control over the audio generation process. Existing methods predominantly rely on global video information, which is insufficient for customizing audio for specific objects or regions within the video according to user needs \cite{2021specvqgan,2024diff-foley,2024SAH}. This hinders the ability of generated audio to meet specific requirements, restricting the practical applicability of V2A models in domains like interactive media and film production. Finer control mechanisms would enhance the accuracy of generated audio and improve the feasibility of V2A models in real-world applications.

To address these challenges, this work proposes an interactive V2A generation framework named Hear-Your-Click. As shown in Fig. \ref{fig:overview}, by allowing users to select specific objects within a video with a single click, Hear-Your-Click generates sounds that correspond to the selected regions, ensuring precise synchronization and alignment with the visual content. This approach provides users with greater control over audio generation, enabling more detailed and customizable interactions with the generated audio. Furthermore, it addresses the suboptimal performance of global V2A approaches with multi-object videos by transferring control to the user, enhancing the quality and applicability of the generated audio.

Our main contributions are:
\begin{itemize} 
\item We propose Hear-Your-Click, an interactive V2A framework that allows users to generate object-specific sounds via a simple click on the videos.

\item To achieve interactive control, we propose Object-aware Contrastive Audio-Visual Fine-tuning (OCAV) alongside Mask-guided Visual Encoder, Random Video Stitching (RVS) and Mask-guided Loudness Modulation (MLM), collectively enhancing the model's responsiveness to selected objects.

\item We conduct thorough experiments to validate our approach on the VGG-AnimSeg dataset which is tailored specifically for our framework. The results across multiple metrics, including our CAV score, showcase state-of-the-art performance in object-level audio-visual alignment.
\end{itemize}
%-------------------------------------------------------------------------

\raggedbottom
\section{Related Work}
\label{sec:formatting}

% {\bf Video-to-Audio Generation.}
\subsection{Video-to-Audio Generation}
Significant advancements in Video-to-Audio (V2A) generation have been fueled by innovations in generative models including Generative Adversarial Networks (GANs) \cite{2020GAN} and diffusion models \cite{2022latentdiffusion}. Various studies propose innovative solutions to V2A challenges. SpecVQGAN \cite{2021specvqgan} trains a codebook on spectrograms to generate sounds. Im2Wav \cite{2023im2wav} employs CLIP embeddings \cite{2021clip} and Transformers \cite{attention} to generate audio from images. Diff-Foley \cite{2024diff-foley} introduces Contrastive Audio-Visual Pre-Training (CAVP) to enhance audio-visual synchronization. Seeing and Hearing \cite{2024SAH} introduces a multimodal latent aligner to bridge existing video and audio generation models. TiVA \cite{2024tiva} and MaskVAT \cite{2024maskvat} focus on synchronization and propose the use of audio layout and sequence-to-sequence mask generative model, respectively. Other methods, such as SonicVisionLM \cite{2024sonicvlm}, SVA \cite{2024sva} and FoleyCrafter \cite{2024foleycrafter}, use text descriptions as a mediating modality between video and audio, allowing them to generate high-quality audio using large language models (LLMs) or pre-trained text-to-audio (T2A) models, but they also face the challenge of achieving accurate synchronization \cite{2024foleycrafter}. The methods above primarily focus on global video information, which can lead to missed local details and difficulties processing complex multi-object videos. Furthermore, they generally lack mechanisms for fine-grained control over the V2A generation process.

%\noindent{\bf Audio-Visual Pre-Training.}
\subsection{Audio-Visual Alignment}
Audio-Visual Pre-Training aims to obtain joint representations for improved retrieval, classification and generation. Methods such as CAV-MAE \cite{2022cav-mae}, AV-MAE \cite{2023av-mae}, MAViL\cite{2024mavil} and CrossMAE \cite{2024crossmae} leverage the Masked Auto-Encoder (MAE) \cite{2022mae} to learn cross-modal correlations, thereby enhancing the performance of joint representations in audio-visual retrieval and classification tasks. Inspired by CLIP \cite{2021clip}, methods like Morgado \cite{2021morgado}, Diff-Foley \cite{2024diff-foley} and SCAV \cite{2024scav} make use of contrastive learning to align audio and video features. Other methods, such as AudioCLIP \cite{2022audioclip} and ImageBind \cite{2023imagebind}, extend multimodal alignment to achieve state-of-the-art results in zero-shot cross-modal tasks. These studies demonstrate the superiority of audio-visual pre-training, and thus an increasing number of works \cite{2024diff-foley,2024SAH} have applied these techniques to V2A generation tasks.

\raggedbottom
\section{Hear-Your-Click}

\subsection{Task Formulation}

Given a $T$-frame video $\mathcal{V}\in\mathbb{R}^{T\times H\times W\times 3}$ and a target object $\mathcal{S}$, our objective is to generate the corresponding Mel spectrogram of the audio $\mathcal{A}$, where $\mathcal{A}\in\mathbb{R}^{T’\times N}$, with $T'$ representing the temporal length and $N$ denoting the mel bins. To achieve this, we first obtain binary masks $\mathcal{M}$ to delineate $\mathcal{S}$, where $\mathcal{M}=\{\mathcal{M}_1,\mathcal{M}_2,\ldots,\mathcal{M}_T\}$ and $\mathcal{M}_t\in\{0,1\}^{H\times W}$ refers to the $t$-th frame mask. In the inference stage, the target object $\mathcal{S}$ is specified through user clicks on video frames. In the training stage, $\mathcal{S}$ is specified through textual prompts labeled by humans. Leveraging the information of $\mathcal{S}$, we perform prompt-guided video segmentation to obtain $\mathcal{M}$. Then $\mathcal{M}$ together with corresponding $\mathcal{V}$ and $\mathcal{A}$ form triplets $\mathcal{(V,M,A)}$, which are used to supervise the model in learning the mappings between the input $(\mathcal{V}, \mathcal{M})$ and the output $\mathcal{A}$. 

\noindent{\bf Method overview.} First, we tailor a new dataset named VGG-AnimSeg based on the task defined above (Sec. \ref{vgganimseg}). Then we propose OCAV to align video-mask-audio triplets (Sec. \ref{racavp}), including a Mask-guided Visual Encoder (MVE) (Sec. \ref{mve}) designed to extract object-level visual features, and two data augmentation strategies: Random Video Stitching (RVS) (Sec. \ref{rvs}) and Mask-guided Loudness Modulation (MLM) (Sec. \ref{mlm}). Finally, we detail the training of a latent diffusion model conditioned on the visual features obtained through OCAV (Sec. \ref{ldm}), along with the construction of the interactive inference framework (Sec. \ref{demo}).

\begin{figure*}[htbp]
	\includegraphics[scale=0.85]{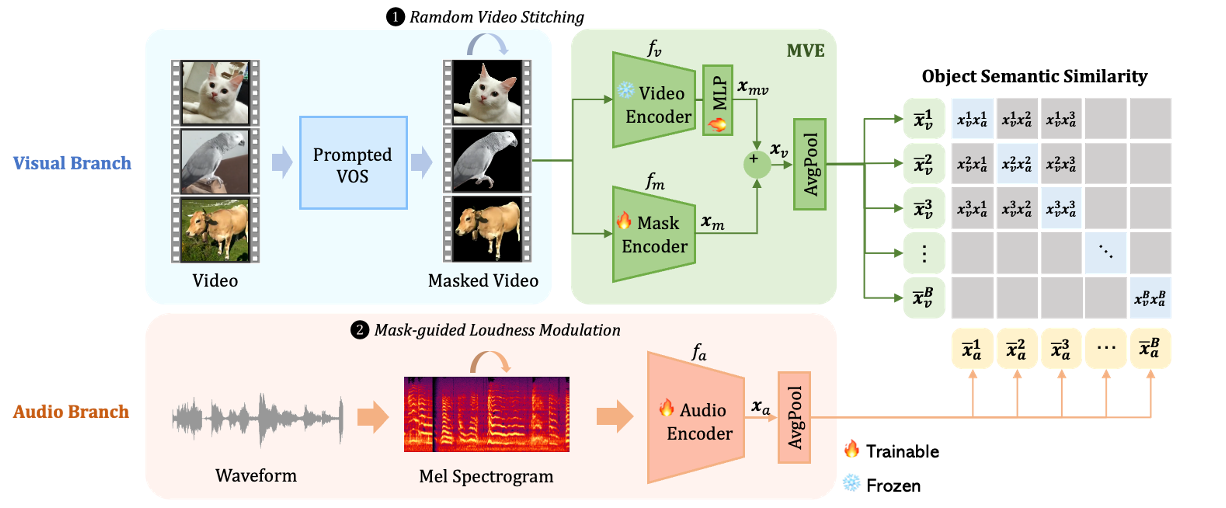}
	\caption{OCAV architecture overview, showing the alignment of object-level visual features from MVE with corresponding audio features. Training data augmentation includes random video stitching and mask-guided loudness modulation.} 
	\label{fig:oacavp}
\end{figure*}

\subsection{VGG-AnimSeg Dataset}
\label{vgganimseg}
Given that existing audio-visual datasets fail to adequately support our objective of focusing on specific objects, we have constructed a new dataset based on the VGGSound dataset \cite{vggsound} to address this limitation. The VGGSound dataset contains over 200K 10-second video clips spanning more than 300 categories, each accompanied by human-labeled textual descriptions. To ensure distinct vocal subjects and cleaner audio, we selected all animal-related videos from VGGSound. This resulted in a subset containing 68 classes of textual descriptions.

To filter out noisy samples, we use multimodal joint embeddings to select samples where audio and video align strongly with textual descriptions. Specifically, we first leverage the Contrastive Language-Audio Pretraining (CLAP) model \cite{2023clap} and Contrastive Language-Image Pre-Training (CLIP) model \cite{2021clip} to extract audio, image, and text embeddings. For each sample, we compute the cosine similarity between the audio-text and image-text embedding pairs. Based on these scores, we select 400 training samples and 40 test samples per textual description with the highest average similarity, resulting in a dataset of approximately 30,000 samples.

To further ensure precise correspondence between audio and visual content, we employ DEVA \cite{deva} to generate video binary masks via text-prompted segmentation using the text descriptions. This ensures a one-to-one correspondence between audio and corresponding visual segments.

\begin{figure*}[ht]
	\centering
	\begin{minipage}[t]{0.48\textwidth}
		\centering
		\includegraphics[width=\linewidth]{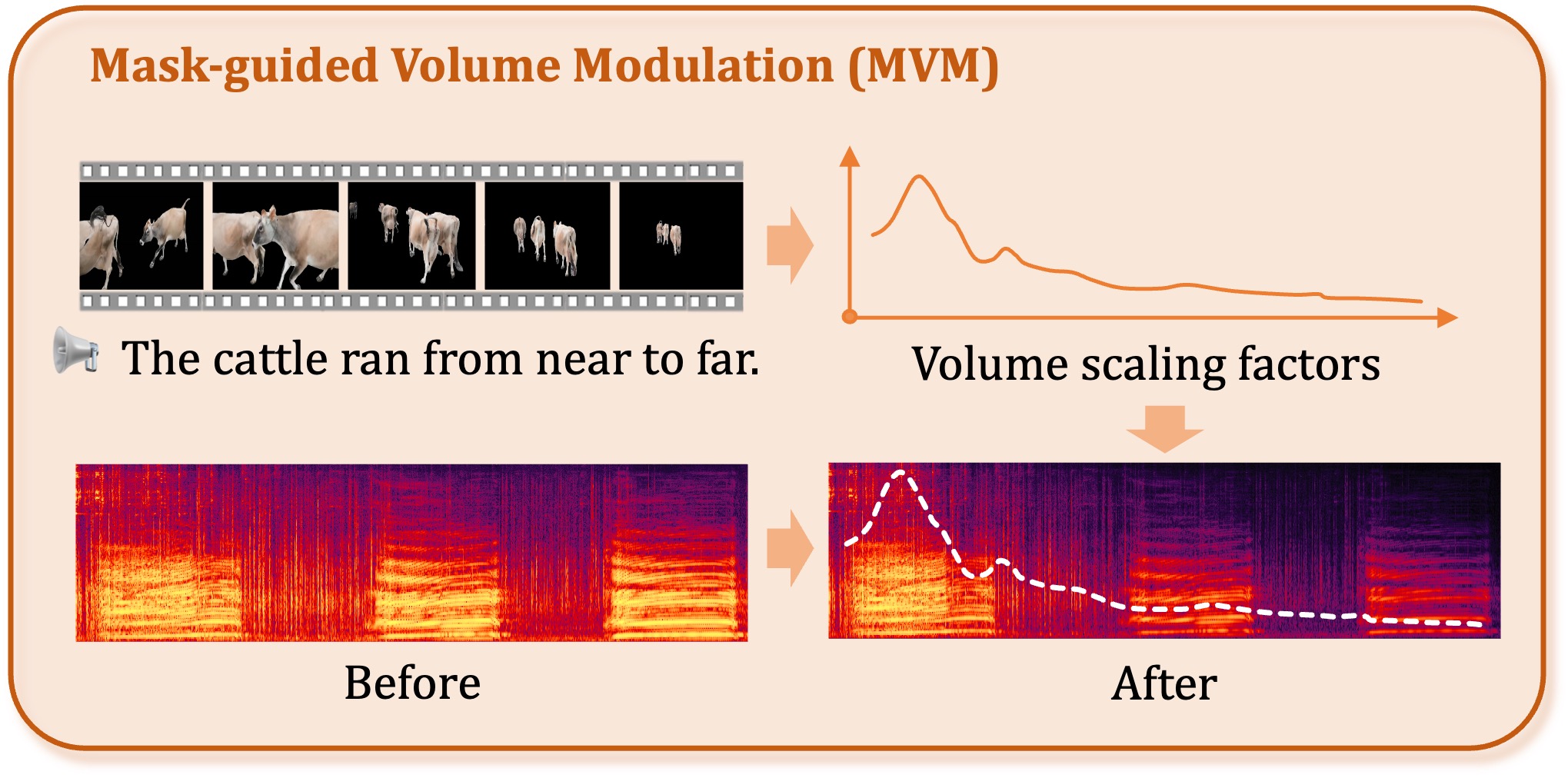}
		\caption{Correlation enhancement between object distance and audio volume changes achieved through adaptive filtering.} 
		\label{fig:mlm_new}
	\end{minipage}
	\hfill
	\begin{minipage}[t]{0.48\textwidth}
		\centering
		\includegraphics[width=\linewidth]{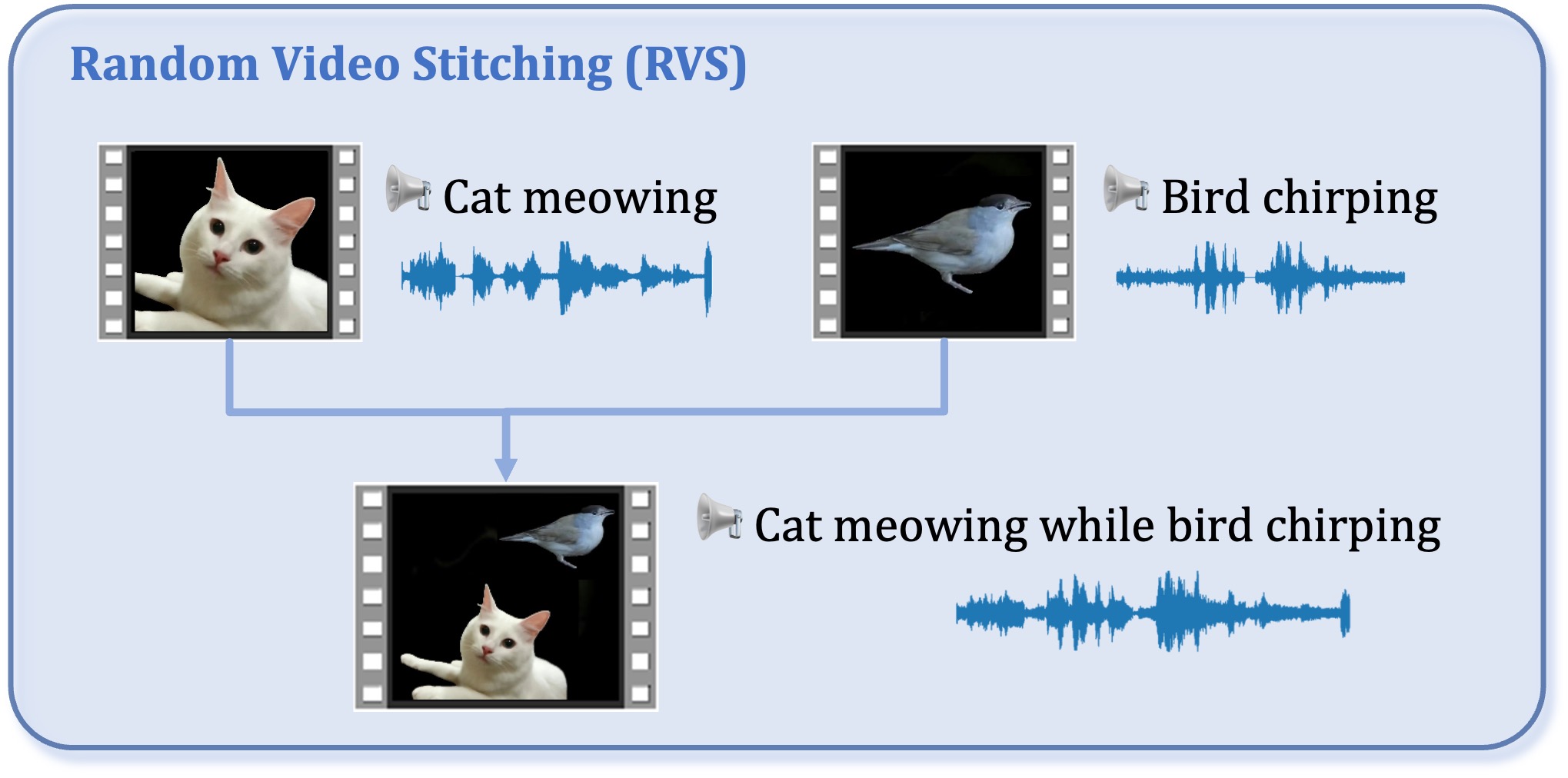}
		\caption{Illustration of the improved multi-object scene handling capability of the model, achieved through a novel attention mechanism focusing on individual objects.} 
		\label{fig:rvs_new}
	\end{minipage}
\end{figure*}

\subsection{OCAV}
\label{racavp}

Audio-visual alignment bridges semantic and temporal gaps between audio and visual modalities, with Contrastive Audio-Visual Pre-training (CAVP) \cite{2024diff-foley} representing a notable advancement in achieving global alignment between these modalities. However, CAVP's focus on global synchronization overlooks fine-grained, object-level details critical for precise audio-visual understanding. To address this limitation while retaining CAVP's strengths, we propose the OCAV framework.

Given triplets $\mathcal{(V,M,A)}$, the objective of OCAV is to extract object-level visual features from video-mask pairs $\mathcal{(V,M)}$, where target objects are determined by $\mathcal{M}$, while ensuring that these visual features are aligned with the features of the audio $\mathcal{A}$. The overview of OCAV is illustrated in Fig. \ref{fig:oacavp}.

\subsubsection{Mask-guided Visual Encoder (MVE)}
\label{mve}

Traditional video encoders, which process the entire video input, often fail to concentrate exclusively on the target objects indicated by the masks $\mathcal{M}$. Therefore, we propose MVE to extract object-specific visual features from video-mask pairs efficiently. The MVE architecture integrates binary masks as supplementary inputs with the primary video stream, enhancing the model's responsiveness to designated objects. This dual-input structure consists of two branches: a video-encoding branch $f_{v}$ and a mask-encoding branch $f_m$. 

When processing the video input for $f_v$, we compared feeding masked videos with original videos and observed that masked videos led to more stable and clearer audio by blocking out background interference. This is denoted as: 
\begin{equation}
	\mathbf{\boldsymbol x}_{mv}=\mathit{norm}(f_{v}(\mathcal{V\odot M}))
\end{equation}
where $\mathbf{\boldsymbol x}_{mv}\in\mathbb{R}^{T\times d}$, $d$ represents the dimension of the features, and $\mathit{norm}(\cdot)$ represents the Euclidean norm along the second dimension to unify the feature scale.

For the mask input, we design $f_m$ with a convolutional backbone to capture the temporal characteristics of the masks. These extracted features are then fused with $\mathbf{\boldsymbol x}_{mv}$ to obtain the ultimate visual features $\mathbf{\boldsymbol x}_{v}$:
\begin{equation}
	\mathbf{\boldsymbol x}_m=\mathit{norm}(f_{m}(\mathcal{M}))
\end{equation}
\begin{equation}
	\mathbf{\boldsymbol x}_{v}=\mathit{norm}(\mathbf{\boldsymbol x}_{mv}+\mathbf{\boldsymbol x}_m)
\end{equation}
where $\mathbf{\boldsymbol x}_m, \mathbf{\boldsymbol x}_{v}\in\mathbb{R}^{T\times d}$. 

\subsubsection{Contrastive Learning}
In parallel with MVE, we adopt a convolutional encoder $f_a$ to extract temporal audio features:
\begin{equation}
\mathbf{\boldsymbol x}_a=\mathit{norm}(f_a(\mathcal{A}))
\end{equation}
where $\mathbf{\boldsymbol x}_a\in\mathbb{R}^{T\times d}$. We then compute the average of $\mathbf{\boldsymbol x}_{v}$ and $\mathbf{\boldsymbol x}_a$ along the time axis to obtain  $\bar{\mathbf{\boldsymbol x}}_v,\bar{\mathbf{\boldsymbol x}}_a\in\mathbb{R}^{d}$.

In the training stage, $f_{v}$ and $f_a$ are initialized with their pre-trained weights from Diff-Foley \cite{2024diff-foley}, whereas $f_m$ is initialized randomly. We freeze most of $f_v$ to retain its prior knowledge, while keeping its final multi-layer perceptron (MLP) block trainable. Both $f_m$ and $f_a$ are kept fully trainable. For each batch, we randomly sample time-synchronized clips from $\mathcal{(V,M,A)}$ and extract feature pairs $\{(\bar{\mathbf{\boldsymbol x}}_v^i,\bar{\mathbf{\boldsymbol x}}_a^i)\}_{i=1}^{B}$, where $B$ is batch size. We use the following contrastive objective to supervise the model training \cite{2021clip,2024LearningTS,2024diff-foley}:
\begin{equation}
	\begin{aligned}
		\mathcal{L}_{\text {contrast}}=\frac{1}{B} \sum_{i=1}^{B} &\left\{-\frac{1}{2}\log \frac{\exp({\phi(\bar{\mathbf{\boldsymbol x}}_v^i, \bar{\mathbf{\boldsymbol x}}_a^i) / \tau})}{\sum_j \exp({\phi(\bar{\mathbf{\boldsymbol x}}_v^i, \bar{\mathbf{\boldsymbol x}}_a^j) / \tau})}\right.\\
		&\left. -\frac{1}{2}\log \frac{\exp  ({\phi(\bar{\mathbf{\boldsymbol x}}_v^i, \bar{\mathbf{\boldsymbol x}}_a^i) / \tau})}{\sum_j \exp ({\phi(\bar{\mathbf{\boldsymbol x}}_v^j, \bar{\mathbf{\boldsymbol x}}_a^i) / \tau})}\right\}
	\end{aligned}
\end{equation}
where $\tau$ is the temperature parameter and $\phi$ denotes cosine similarity function:

\begin{equation}
\phi (\mathbf{\boldsymbol a}, \mathbf{\boldsymbol b})=\frac{\mathbf{\boldsymbol a} \cdot \mathbf{\boldsymbol b}}{\|\mathbf{\boldsymbol a}\|\|\mathbf{\boldsymbol b}\|}
\end{equation}

\subsubsection{Mask-guided Loudness Modulation (MLM)}
\label{mlm}
Temporal inconsistencies in the dataset, such as sounds continuing after their sources move out of the frame, create ambiguities for the models. These ambiguities make it difficult for the models to determine whether sounds should continue or stop when the sound sources are no longer visible. Imperfect audio-video synchronization, including off-screen sounds and background noise, complicates audio-video alignment and introduces interference to joint representation learning.

We aim to mitigate these issues by ensuring the cessation of sound effects precisely corresponds to the target object's departure from the camera view during inference. Additionally, we aim to adjust audio loudness according to the objects' relative distances from the camera. To accomplish this, we incorporate Mask-guided Loudness Modulation (MLM) into our training methodology. By strengthening the correlation between object masks and audio loudness, MLM facilitates more accurate learning of the mapping between visual masks and auditory components.

As illustrated in Fig. \ref{fig:oacavp}, given a set of binary masks $M$, we first compute the ratio of unmasked pixels to the total number of pixels for each frame:
\begin{equation}
	\lambda_k=\frac{\sum_{\substack{0 \leq i<W \\ 0 \leq j<H}} \mathcal{M}_k(i, j)}{H \times W}, k=1,2, \ldots, T
\end{equation}

\noindent Then we normalize these values to ensure the maximum value is 1:
\begin{equation}
	\lambda_k^{\prime}=\frac{\lambda_k}{\max \left(\lambda_1, \lambda_2, \ldots, \lambda_T\right)}, k=1,2, \ldots, T
\end{equation}

\noindent Next, we apply linear interpolation to resample these values from the length $T$ to match the length of 1-D audio signals, thereby obtaining a new set of loudness scaling factors $\Lambda=\{\lambda'_1, \lambda'_2, \ldots, \lambda'_T\}$. Finally, $\Lambda$ is element-wise multiplied with audio signals to modulate the loudness based on the presence and extent of the unmasked region over time.

\subsubsection{Random Video Stitching (RVS)}
\label{rvs}
Video augmentation methods are extensively applied in various video processing tasks, including classification, action recognition and object detection \cite{yun2019cutmix,chen2023spatial,duan2023improve,xu2024shortform}. These techniques facilitate more robust generalization and improve the performance of deep learning models by generating diverse training data. Moreover, studies such as LeMDA \cite{2022ledma} and MixGen \cite{2023mixgen} offer novel insights into multimodal data augmentation. However, such methods have been underutilized in the domain of audio-visual alignment.

We propose RVS during OCAV to enhance the model's ability to identify and understand individual objects. Specifically, for each video in the dataset, another video is randomly selected, and each frame is stitched either horizontally or vertically with the corresponding frame from the original video. Simultaneously, the audio from both videos is overlapped. By incorporating RVS, we aim to improve the model's capacity to handle complex, multi-object scenes and enhance its performance and generalization.

Theoretically, RVS can significantly increase training samples by combining different video segments; however, too many augmented samples can lead to overfitting and reduced performance on new data. Therefore, we apply RVS to the 1,126 training samples from the VGG-AnimSeg-1k dataset and integrate the augmented data into the final training set to ensure a balance between original and augmented samples.

\begin{figure*}[htbp]
	\centering
	\includegraphics[scale=0.11]{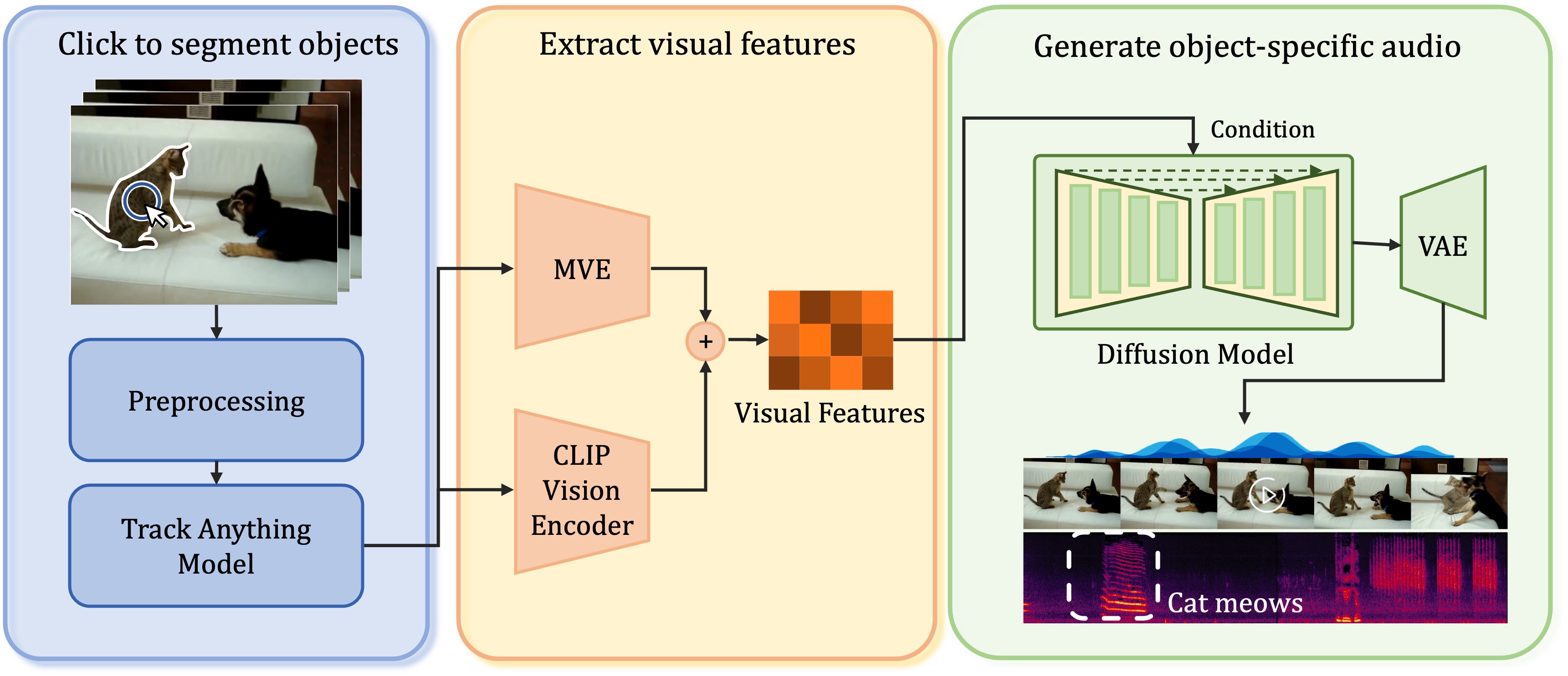}
	\caption{\textbf{Overview of interactive inference process.} Users can upload a silent video and select a frame to refine the mask via clicks based on Segment Anything Model (SAM). The Track Anything Model (TAM) then propagates these masks for MVE and CLIP to extract visual features. Finally, a trained LDM generates the final audio conditioned on these features.} 
	\label{fig:inference}
\end{figure*}

\subsection{Latent Diffusion Model}
\label{ldm}
We utilize Latent Diffusion Models (LDMs) \cite{2022latentdiffusion} as our generative model due to their proven efficacy in cross-modal generation tasks. To mitigate the issues of failing to generate audio for the target object or generating incorrect audio, we propose augmenting the MVE visual features $\boldsymbol x_v$ with CLIP features $\boldsymbol x^{*}_v$ of the masked frames, as they provide rich, high-level semantic information that helps disambiguate between different objects. This is denoted as:
\begin{equation}
	\boldsymbol x_{c}=\boldsymbol x_v+\boldsymbol x^{*}_v
\end{equation}
where $\boldsymbol x_{c}, \boldsymbol x^{*}_v\in\mathbb{R}^{T\times d}$, and $\boldsymbol x^{*}_v$ is extracted from the T frames of a video. Then we train our model conditioned on $\boldsymbol x_{c}$.

Given a Mel spectrogram $\mathcal{A}\in\mathbb{R}^{T'\times N}$, we first map it into a latent space through $\boldsymbol z_0=\boldsymbol E(\mathcal{A})$, where $\boldsymbol z_0\in \mathbb{R}^{C\times T^{*}\times N^{*}}$ and $\boldsymbol E(\cdot)$ is a pre-trained latent encoder. In the latent space, LDM employs a forward diffusion process to gradually add noise to $\boldsymbol z_0$, transforming it to a Gaussian distribution $\mathcal{N}$:
\begin{equation}
	q(\mathbf{\boldsymbol z}_t \mid\mathbf{\boldsymbol z}_{t-1})=\mathcal{N}(\boldsymbol z_t ; \sqrt{1-\beta_t} \boldsymbol z_{t-1}, \beta_t \mathbf{I})
\end{equation}
\begin{equation}
	q(\boldsymbol z_{t} \mid \boldsymbol z_{0})=\mathcal{N}(\boldsymbol z_{t} ; \sqrt{\bar{\alpha}_{t}} \boldsymbol z_{0},\left(1-\bar{\alpha}_{t}\right) \mathbf{I})
\end{equation}
where $\bar{\alpha}_{t}=\prod_{i=1}^{t}\left(1-\beta_{i}\right)$, $\beta_t$ is the noise schedule parameter that controls the amount of noise added at each time step. Then LDM is trained to predict the noise added at each step of the forward diffusion process:
\begin{equation}
	\mathcal{L}_{L D M}=\mathbb{E}_{\boldsymbol z_{0}, t, \boldsymbol \epsilon}\left\|\boldsymbol \epsilon-\boldsymbol \epsilon_{\theta}\left(\boldsymbol z_{t}, t, \boldsymbol x_c\right)\right\|_{2}^{2}
\end{equation}
where $\boldsymbol \epsilon\sim\mathcal{N}(0,\mathbf{I})$ and $\boldsymbol \epsilon_{\theta}$ is predicted by the model. After training, LDM follows the reverse process to generate new samples. Starting from a random noise latent $z_{T}\sim \mathcal{N}(0,\mathbf{I})$, the model iteratively denoises to reconstruct a meaningful latent:
\begin{equation}
	p_{\theta}(\boldsymbol z_{t-1} \mid \boldsymbol z_{t},\boldsymbol x_c)=\mathcal{N}(\boldsymbol z_{t-1} ; \mu_{\theta}(\boldsymbol z_{t}, t, \boldsymbol x_c), \sigma_{t}^{2} \mathbf{I})
\end{equation}
\begin{equation}
	p_{\theta}(\boldsymbol{z}_{0: T} \mid \boldsymbol x_c)=p\left(\boldsymbol{z}_{T}\right) \prod_{t=1}^{T} p_{\theta}(\boldsymbol{z}_{t-1} \mid \boldsymbol{z}_{t}, \boldsymbol x_c)
\end{equation}
where $\mu_{\theta}$ and $\sigma_{t}^{2}$ are defined as follows:
\begin{equation}
	\mu_{\theta}(\boldsymbol z_{t}, t, \boldsymbol x_c)=\frac{1}{\sqrt{\alpha_{t}}}(\boldsymbol z_{t}-\frac{1-\alpha_{t}}{\sqrt{1-\bar{\alpha}_{t}}} \boldsymbol \epsilon_{\theta}(\boldsymbol z_{t}, t, \boldsymbol x_c))
\end{equation}
\begin{equation}
		\sigma_{t}^{2}=\frac{1-\bar{\alpha}_{t-1}}{1-\bar{\alpha}_{t}}(1-\alpha_{t})
\end{equation}
$\boldsymbol z_0$ is then mapped back to the original data space using the decoder $\boldsymbol{D}$: $\hat{\mathcal{A}}=\boldsymbol{D}(\boldsymbol z_0)$.

\subsection{Interactive Inference}
\label{demo}
There is a growing trend in research towards interactive content generation \cite{2023tokenflow,2024instancediffusion,2024follow-your-click,2024generative} because it engages users directly in the process, allowing for real-time feedback and customization. This makes the system more adaptable to specific user needs and enhances the user experience. Inspired by these studies, we develop an interactive interface for Hear-Your-Click, utilizing the promptable segmentation capabilities of the Segment Anything Model (SAM) \cite{sam} and the Track Anything Model (TAM) \cite{trackanything}. This interface allows users to upload a silent video and then select a single frame to specify target regions via clicks. Upon selection, SAM generates corresponding masks in real time based on user clicks. Users can iteratively refine these masks to ensure precise delineation of the target object. Once the mask is finalized, TAM propagates it throughout the entire video sequence using semi-supervised video object segmentation, generating the comprehensive video mask $\mathcal{M}$. Next, we extract MVE features $\boldsymbol x_v$ and CLIP features $\boldsymbol x_v^{*}$ from $\mathcal{(V,M)}$, forming the condition embedding $\boldsymbol x_c$. Finally, we sample the audio using the trained LDM conditioned on $x_c$. Fig. \ref{fig:inference} provides a detailed overview of the inference pipeline, from user interaction to audio generation.

\raggedbottom
\section{Experiments}

\subsection{Experimental Setup}
\label{experimentalsetup}

%\noindent{\bf \repl{VGG-AnimSeg} Dataset\pdfcomment[color=orange]{Original: Dataset. Explanation: To make it clear which dataset this section refers to, it's best to include the specific dataset name, especially since you have dedicated a whole section to the creation of this dataset.}s.}
\subsubsection{Dataset}

Our models are trained and evaluated on the VGG-AnimSeg dataset, consisting of 27,200 training samples and 2,720 testing samples. When data augmentation via Random Video Stitching (RVS) is applied, an additional 6,800 augmented training samples are incorporated, resulting in 34,000 samples for model training.

% \noindent{\bf Evaluation Metrics.}
\subsubsection{Evaluation Metrics}

For quantitative assessment, we use Frechet Distance (FD), Frechet Audio Distance (FAD), Inception Score (IS), Kullback-Leibler Divergence (KL) and Kernel Inception Distance (KID) following other audio generation studies \cite{2021specvqgan,2023audioldm,2024audioldm2,2024diff-foley,2024SAH}. FD, FAD, KL and KID are employed to quantify the similarity between the generated audio and the ground truth, while IS is utilized to assess the diversity and quality of the generated audio samples. To better evaluate the correspondence between video and audio, we introduce a new metric, the CAV score. The score leverages the model from C-MCR \cite{cmcr}, which integrates CLIP \cite{2021clip} and CLAP \cite{2023clap} to provide audio-image contrastive representations. For each audio-video pair, we obtain per-frame image embeddings and an audio embedding, and then calculate the similarity between the average image embedding and the audio embedding. A higher CAV score indicates a better match between the generated audio and the original video.

\begin{figure*}[t]
	\centering
	\begin{subfigure}[t]{0.5\linewidth}
		\captionsetup{width=0.92\linewidth}
		\centering
		\includegraphics[scale=0.45]{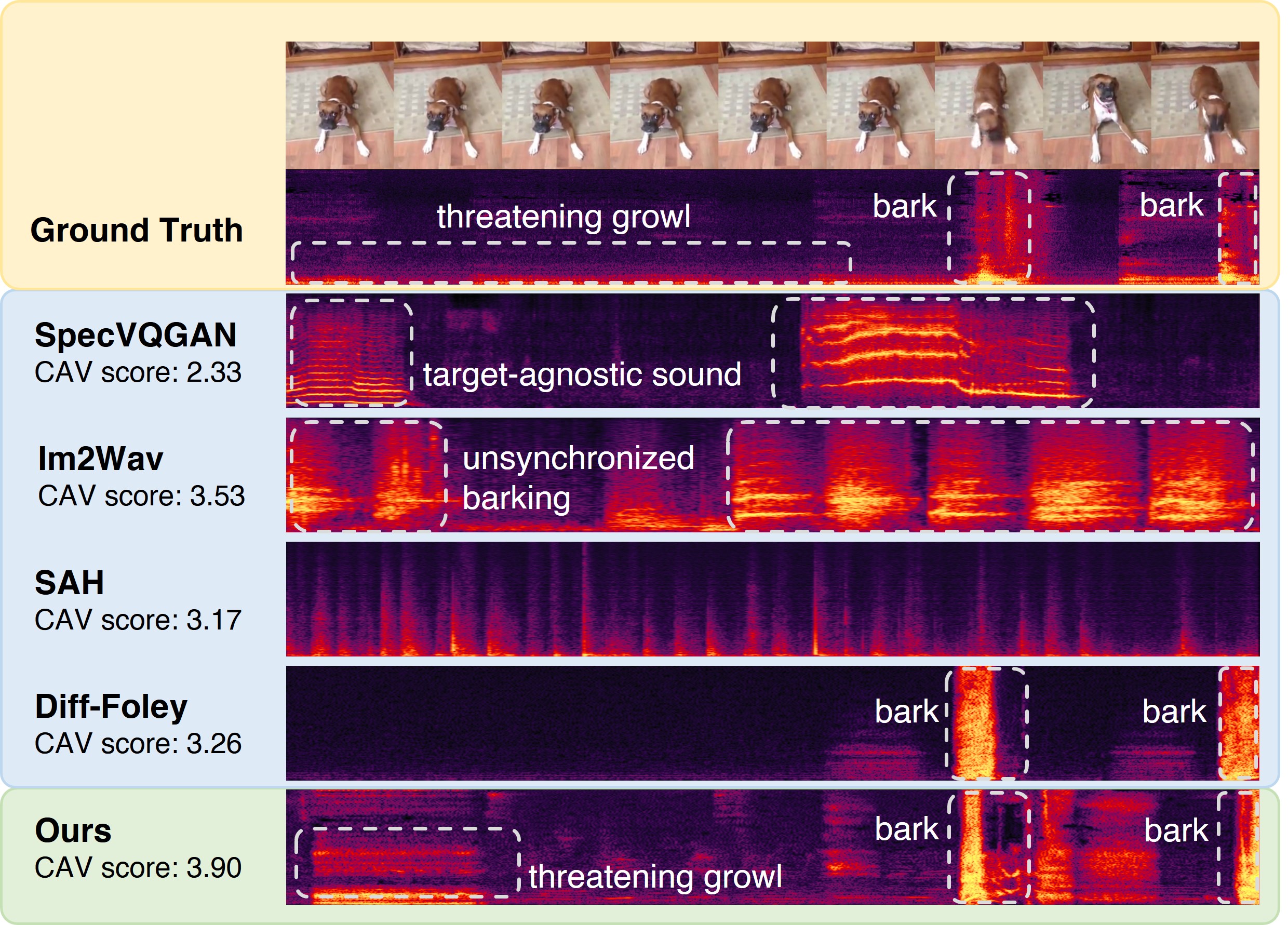}
		\caption{The dog growls and suddenly barks loudly.}
		\label{fig:result-a}
	\end{subfigure}%
	\begin{subfigure}[t]{0.5\linewidth}
		\captionsetup{width=0.92\linewidth}
		\centering
		\includegraphics[scale=0.45]{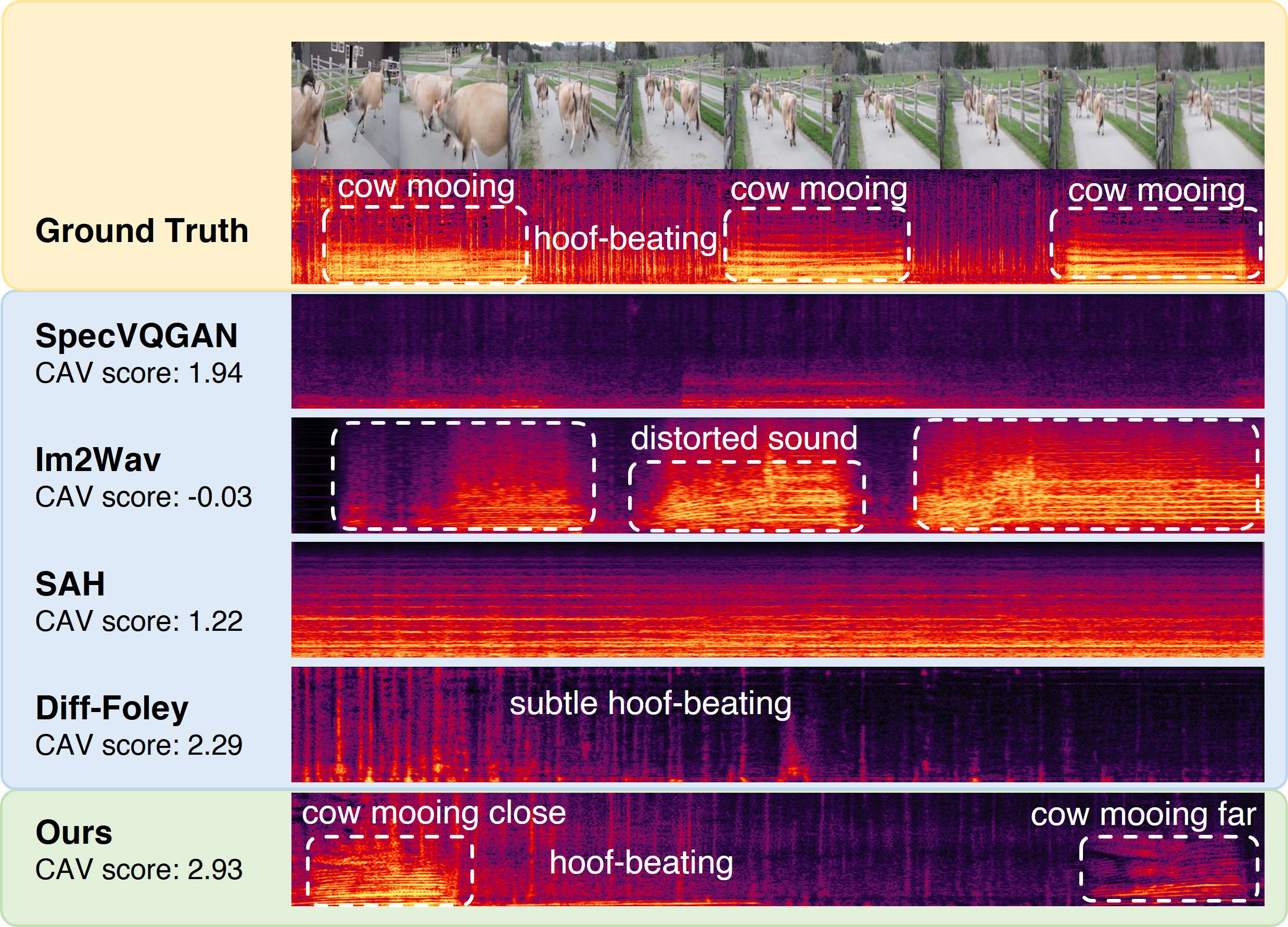}
		\caption{Herds of cattle pass by.}
		\label{fig:result-b}
	\end{subfigure}
	\caption{Visualization of generated audio samples: (a) Reproduction of ground truth sounds with preserved synchronization, demonstrating accurate capture of transient motion. (b) Audio volume reduction corresponding to cows running away, while maintaining finer details.}
	\label{fig:singleresult}
\end{figure*}

%\noindent{\bf Implementation Details.}
\subsubsection{Implementation Details}

For the VOS setting, the IoU threshold and Non-Maximum Suppression (NMS) threshold of SAM are set to 0.88 and 0.8. TAM is configured to use 15 voting frames. For data preprocessing, we resample 10-second video clips to 4 fps and resize each frame to 224$\times$224 pixels. For audio samples, we compute Mel spectrograms with a hop size of 250 and 128 mel bins during OCAV training, and a hop size of 256 and 128 mel bins during LDM training. Considering the limited size of our dataset, we construct our framework based on Diff-Foley and fine-tune it using pre-trained weights, since training from scratch is insufficient. Specifically, the $f_v$ and $f_m$ in MVE utilize the SlowOnly \cite{slowfast} architecture, known for its proficiency in detecting temporal action changes within videos. $f_a$ leverages an audio encoder provided by PANNs \cite{panns}. The architecture of LDM, including the latent encoder and decoder, is inherited from Stable Diffusion-V1.4 (SD-V1.4) \cite{2022latentdiffusion}. During the OCAV training process, we randomly sample 4-second segments from the original audio and videos, denoted as $\mathcal{A}\in\mathbb{R}^{256\times 128}$ and $\mathcal{V}\in\mathbb{R}^{16\times 224\times 224\times 3}$, yielding corresponding features $\boldsymbol{x}_a\in\mathbb{R}^{16\times 512}$, $\boldsymbol{x}_v\in\mathbb{R}^{16\times 512}$. For LDM training, we first extract visual features through MVE from the 10-second videos, resulting in  $\boldsymbol{x}_v\in\mathbb{R}^{40\times 512}$. Following the Im2Wav \cite{2023im2wav} method, we extract per-frame CLIP embeddings, resulting in $\boldsymbol{x}^{*}_v\in\mathbb{R}^{40\times 512}$, and add them to $\boldsymbol{x}_v$ to form the condition embedding $\boldsymbol{x}_c$ for the LDM. At the inference stage, we set the Classifier-Free Guidance Scale to 4.5, the Classifier-Guidance Scale to 50, and use the DPM-Solver \cite{dpm} sampler with 50 inference steps.

\subsection{Comparative Experiments}
%\noindent{\bf Qualitative results.}
\subsubsection{Qualitative Results}
We conduct a comparative experiment of our method with several recently published, open-source V2A approaches, including SpecVQGAN \cite{2021specvqgan}, Im2Wav \cite{2023im2wav}, Seeing and Hearing (SAH) \cite{2024SAH} and Diff-Foley \cite{2024diff-foley}. The qualitative evaluation focuses on three main aspects: the capacity to produce coherent audio, the relevance of the generated audio to the target objects, and the temporal alignment between audio and visual content, particularly the movements. As shown in Fig. \ref{fig:singleresult}, our method excels in generating audio most similar to the original for given video inputs, accurately capturing nuances such dog barks and cow moos, while maintaining synchronization with their movements. Other methods fall short, either producing visually accurate but poorly synchronized or low-quality audio (Im2Wav) or generating irrelevant sounds (SAH). SpecVQGAN, despite its high spectral detail, fails to correctly identify video objects, resulting in inaccurate audio generation. While Diff-Foley succeeds in crafting synchronized audio, it occasionally struggles to reproduce the exact sounds associated with specific objects. Overall, our approach demonstrates a superior balance between audio quality, synchronization, and object-specific sound accuracy.

\begin{table}[htbp]
	\setlength\tabcolsep{3pt}
	\begin{tabular}{p{2.2cm}@{}lcccccc@{}}
		\toprule
		Method        & FD↓   & FAD↓          & IS↑           & KL↓    & KID↓  & CAV↑ \\ \midrule
		SpecVQGAN     & 75.55 & 5.51          & 3.79          & 3.53   & 0.021 & 1.50       \\
		Im2Wav        & 63.35 & 11.94         & 3.27          & 3.13   & 0.021 & 2.27       \\
		SAH           & 98.41 & 11.30         & 4.46          & 5.09  & 0.039 & 1.76       \\
		Diff-Foley    & 59.00 & 6.60          & \textbf{5.67} & 3.95   & 0.015 & 2.22       \\
		Ours$^1$ & 49.48 & \textbf{4.04} & 5.20          & 3.01 & 0.012 & 2.55       \\
		Ours$^2$ & \textbf{48.78} & 5.02 & 4.49 & \textbf{2.82} & \textbf{0.010} & \textbf{2.67} \\ \bottomrule
	\end{tabular}
	\caption{Results of quantitative comparison experiments. Our$^1$ represents inference conditioned on MVE features, and Our$^2$ represents inference conditioned on MVE+CLIP features.}
	\label{tab:comparison}
\end{table}

%\noindent{\bf Quantative results.}
\subsubsection{Quantitative Results}
For each video from VGG-AnimSeg-4k test set, we generate an 8.2-second audio clip for evaluation. Tab. \ref{tab:comparison} demonstrates our method presents superior performance compared to other methods in terms of FD, FAD, KL, and KID, indicating that our results closely resemble the ground truth. Our excellent CAV score further highlights our method's exceptional alignment between the generated audio and input videos.

\begin{figure}[htbp]
	\includegraphics[scale=0.57]{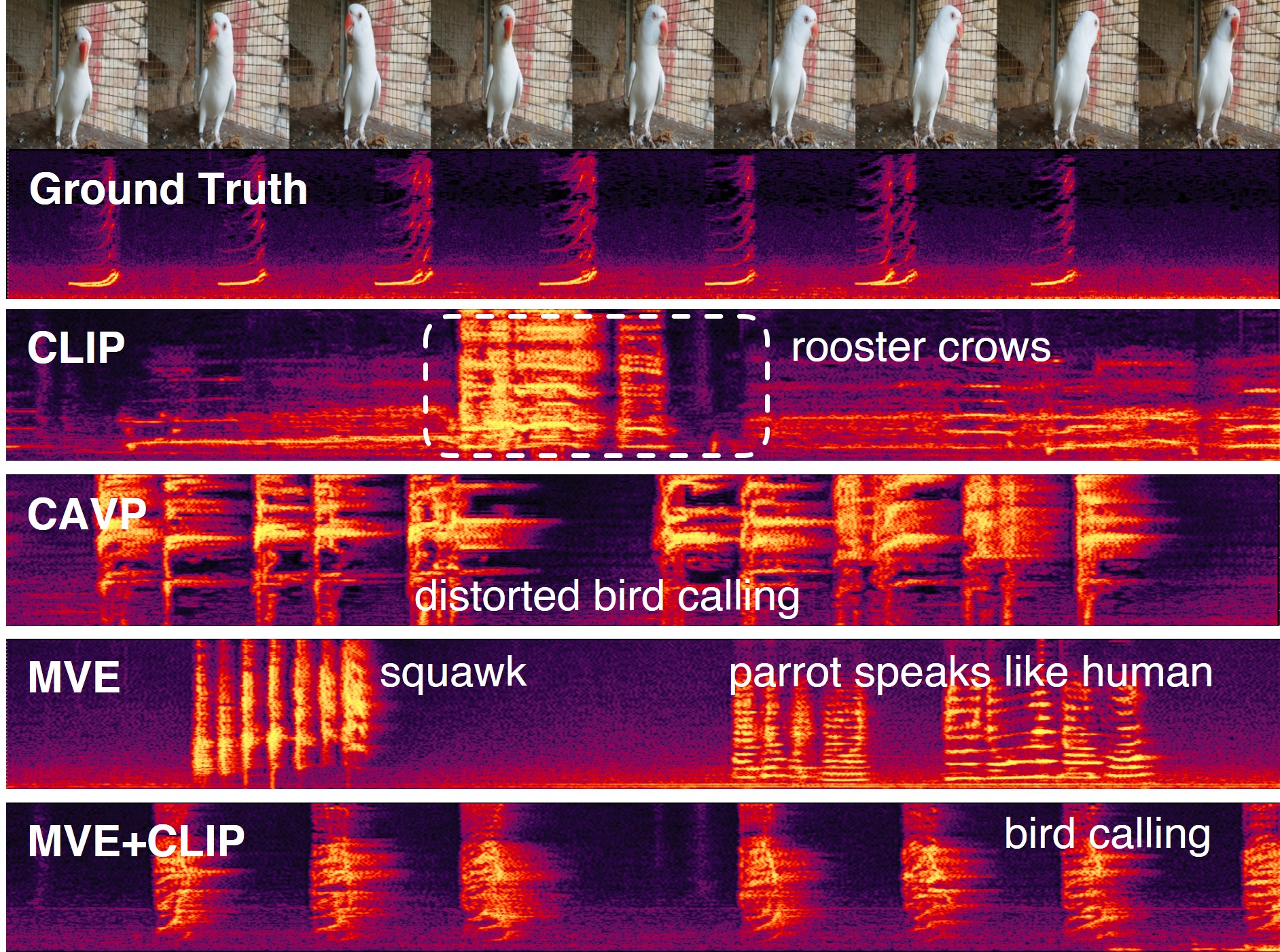}
	\caption{Comparison of audio generation results using different visual features, demonstrating that MVE and MVE+CLIP capture semantic information and temporal context more effectively than other methods, which produce target-agnostic or low-fidelity audio.} 
	\label{fig:visfea}
\end{figure}

\subsection{Ablation Studies}
%\noindent{\bf MVE.}
\subsubsection{MVE Ablation}
To validate the effectiveness of the visual features extracted by MVE, we compare them with other visual features. We fine-tune the LDM using different visual features extracted from the VGG-AnimSeg-4k training set, and then generate 10 audio samples for each test video for evaluation.
As shown in Tab. \ref{tab:mve}, our method achieves best in multiple metrics. Although CAVP features achieve the highest IS score, they perform worse on other metrics. Since IS is originally designed to measure the quality of generated images, it does not guarantee the quality of the generated audio. Additionally, adding CLIP features to MVE features (referred to as MVE+CLIP) further enhances performance. This improvement is attributed to the rich, high-level semantic information provided by CLIP features, which a convolutional backbone may ignore. Fig. \ref{fig:visfea} shows a qualitative comparison of a representative example.

\begin{table}[h]
	\setlength\tabcolsep{3pt}
	\begin{tabular}{p{2.3cm}@{}lccccc@{}c@{}}
		\toprule
		Visual Features & FD↓            & FAD↓            & IS↑           & KL↓           & KID↓           & CAV↑     \\ \midrule
		CLIP            & 42.78          & 5.73           & 5.39          & 3.12          & 0.015          & 2.55           \\
		CAVP            & 47.38          & 6.73           & \textbf{6.65} & 4.14          & 0.016          & 1.95           \\
		MVE         & 38.89          & \textbf{4.03} & 5.79          & 3.17          & 0.012          & \textbf{3.06} \\
		MVE+CLIP    & \textbf{35.41} & 4.90           & 5.93          & \textbf{2.90} & \textbf{0.011} & 2.69           \\ \bottomrule
	\end{tabular}
	\caption{Investigation of the impact of conditioning the model on different visual features, demonstrating that MVE features outperform other methods, and MVE+CLIP features lead to further performance improvements.}
	\label{tab:mve}
\end{table}

%\noindent{\bf MLM and RVS Ablation.}
\subsubsection{MLM and RVS Ablation}
\begin{figure}[htbp]
	\centering
	\includegraphics[scale=0.56]{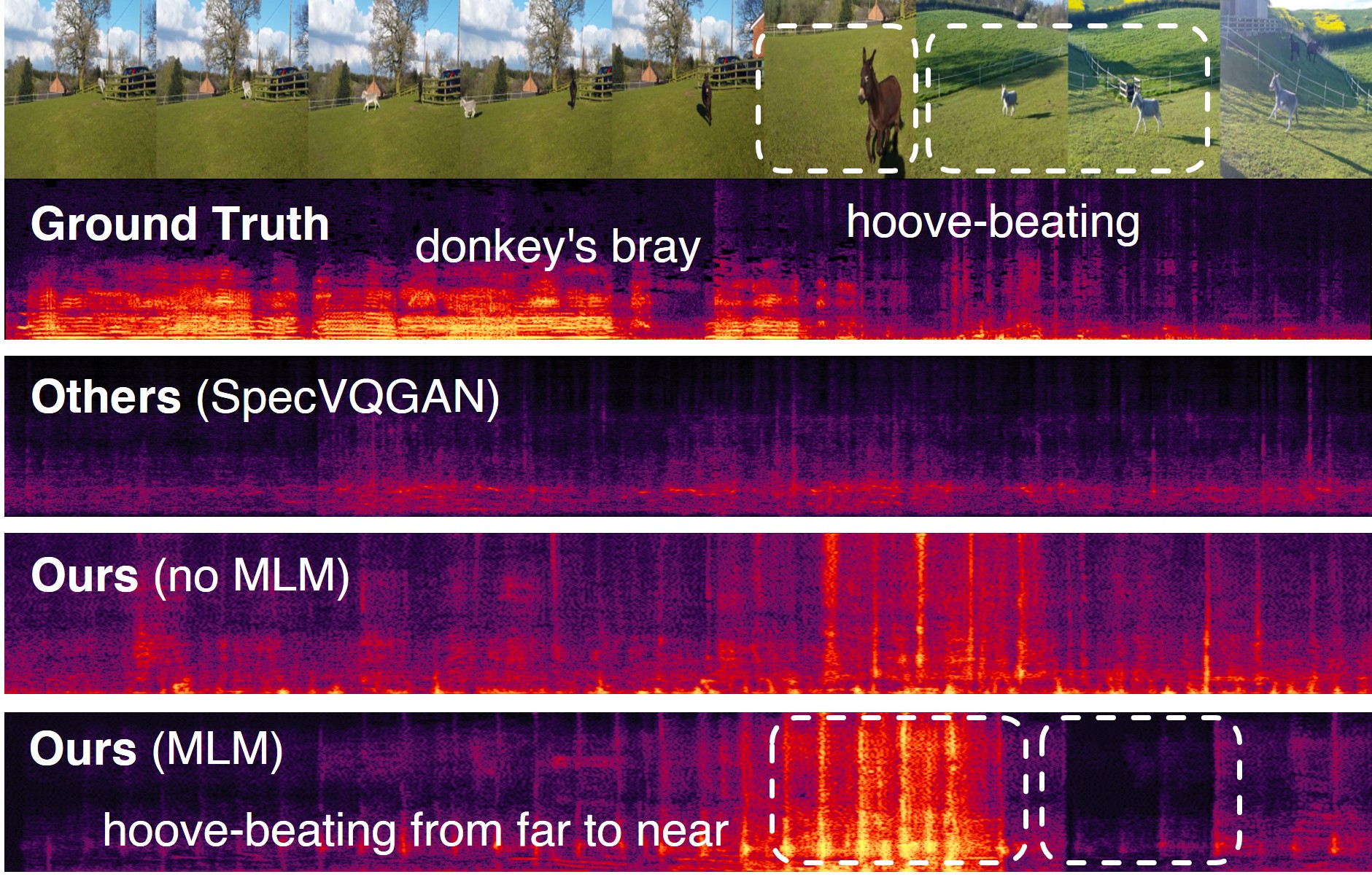}
	\caption{Ablation study of the Masked Loudness Modulation (MLM) component, demonstrating that as the donkey moves closer, the generated audio exhibits a clearer increase in loudness, and diminishes as it moves away.}
	\label{fig:mlm}
\end{figure}

\begin{figure}[htbp]
	\centering
	\includegraphics[scale=0.56]{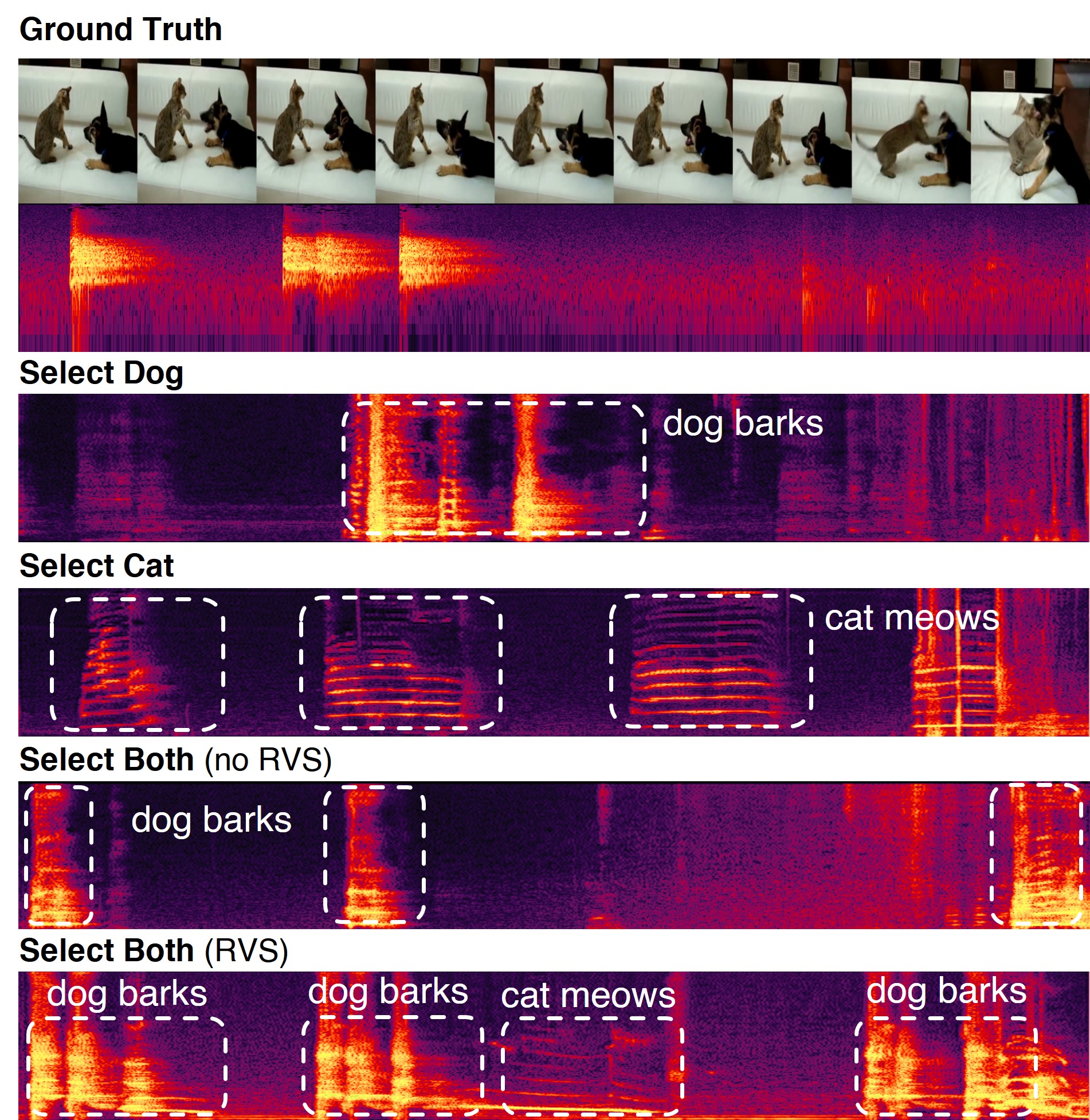}
	\caption{Ablation study of Random Video Stitching (RVS), demonstrating its role in enabling the model to generate distinct sounds corresponding to multiple selected objects in the scene.}
	\label{fig:rvs}
\end{figure}

We conduct an ablation study to validate the effectiveness of RVS and MLM. As shown in Tab. \ref{tab:mlmrvs}, MLM significantly improves the scores of FD, FAD, KL, KID, and CAV. The qualitative comparison in Fig. \ref{fig:mlm} demonstrates that the model trained with MLM generates audio with more pronounced changes in loudness corresponding to the distance of the target object. However, adding RVS to our base model slightly decreases performance. This may be due to the resizing process in RVS, which can distort the aspect ratio of the original images. Despite this, Fig. \ref{fig:rvs} demonstrates that RVS enables the model to handle multi-object scenes more effectively, highlighting its importance in complex scenarios.

\begin{table}[h]
	\setlength\tabcolsep{3pt}
	\begin{tabular}{@{}cccccccc@{}}
		\toprule
		MLM        & RVS        & FD↓            & FAD↓          & IS↑           & KL↓           & KID↓           & CAV↑ \\ \midrule
		&            & 41.29          & 5.22          & \textbf{5.86}          & 3.36          & 0.012          & 2.28   \\
		\checkmark &            & \textbf{34.67} & 4.81          & 5.53          & \textbf{3.07} & \textbf{0.010} & \textbf{2.47}   \\
		& \checkmark & 50.19          & 4.87          & 5.56          & 3.62          & 0.018          & 2.30   \\
		\checkmark & \checkmark & 40.11          & \textbf{4.81} & 5.37 & 3.16          & 0.014          & 2.35   \\ \bottomrule
	\end{tabular}
	\caption{Validation of the Masked Loudness Modulation (MLM) and Random Video Stitching (RVS) components through quantitative metrics.}
	\label{tab:mlmrvs}
\end{table}

\section{Conclusion}
In this work, we introduce Hear-Your-Click, an interactive V2A framework. To support this framework, we develope the VGG-AnimSeg dataset and propose Object-aware Contrastive Audio-Visual Fine-tuning, which includes a Mask-Guided Visual Encoder and two data augmentation strategies. For a more precise evaluation of audio-visual alignment, we introduce the CAV score alongside traditional metrics. Our experimental results demonstrate superior performance across various objective indicators and qualitative assessments. Additionally, we have created an interactive V2A interface for Hear-Your-Click, a feature not found in other V2A methods. These results highlight the effectiveness of our approach in capturing object-specific local details within videos, leading to more accurate audio generation. We anticipate that our work will inspire further advancements in V2A research and facilitate its broader application.

\bibliographystyle{ACM-Reference-Format}
\bibliography{sample-base}

\end{document}